\documentclass[letterpaper, 10 pt, conference]{ieeeconf}  

\IEEEoverridecommandlockouts                              

\usepackage{graphics} 
\usepackage{epsfig} 
\usepackage{amsmath} 
\usepackage{array}
\usepackage{cite}
\usepackage{float}
\usepackage[font=footnotesize,labelsep=space]{caption}
\usepackage{makecell}
\usepackage{multirow}
\usepackage{xcolor}
\usepackage{url}
\def\BibTeX{{\rm B\kern-.05em{\sc i\kern-.025em b}\kern-.08em
    T\kern-.1667em\lower.7ex\hbox{E}\kern-.125emX}}

\title{\LARGE \bf
Modular Pipe Climber III with Three-Output Open Differential
}

\author{\normalsize Rama Vadapalli$^{1}$, Saharsh Agarwal$^{1}$, Vishnu Kumar$^{1}$, Kartik Suryavanshi$^{1}$, Nagamanikandan G$^{1}$ and K Madhava Krishna$^{1}$
\vspace{-0.02in}
\thanks{*This work was not supported by any organization}
\thanks{All the authors are with Robotics Research Center,
        International Institute of Information Technology, Hyderabad, India. 500032
        {\tt\small 
        e-mail: rvvadapalli@gmail.com}}
}

\begin{document}

\maketitle
\thispagestyle{empty}
\pagestyle{empty}
\begin{abstract}
The paper introduces the Modular Pipe Climber III with a novel Three-Output Open Differential mechanism to eliminate slipping of the tracks due to the changing cross-sections of the pipe. This will be achieved in any orientation of the robot. Previously, pipe climbers used three-wheel/track modules with an individual driving mechanism to achieve stable traversing. Slipping of tracks is prevalent in such robots when it encounters the pipe turns. Thus, active control of each module's speeds is employed to mitigate the slip at the expense of control effort. Our Pipe climber robot addresses this issue by using our Three-Output Differential (3-OOD) mechanism that provides embodied intelligence to the robot to modulate the track speeds mechanically as it encounters the turns.


\end{abstract}
\section{INTRODUCTION}
Pipelines are predominantly used in industries for the transportation of gases, oils and various other fluids \cite{poznanski1978implementation}. They require frequent inspection and maintenance to prevent damage due to scaling and corrosion. Due to their inaccessibility, pipeline inspections are often complicated and expensive, for which robotic inspection is a feasible solution \cite{kawaguchi1995internal}, \cite{roman1993pipe}. A wide variety of drive-mechanisms have been explored in the past decades such as wheeled, screw, tracked, pipe inspection gauge, inchworm, articulated and few others \cite{kwon2012design,li2015design,singh2017cocrip,hirose1999design,kim2010famper,roslin2012review,suzumori2003miniature,dertien2014design,hu2005dynamic,kwon2011flat}. However, most of them used multiple actuators and active steering which increased the control efforts to steer and maneuver inside the pipe, causing inaccurate localization due to slip while traversing in bends. Generally, three-tracked in-pipe robots are shown to be dynamically more stable with better mobility \cite{roh2005differential}. Our previously developed robots, the Modular Pipe Climber I \cite{vadapalli2019modular} and the Omnidirectional Tractable Three Module Robot (Modular Pipe Climber II) \cite{suryavanshi2020omnidirectional}, both incorporated three-tracked driving systems with active differential speed to smoothly negotiate bends in pipe networks. The speeds of the three tracks of the robot were predefined in accordance to the bends in the pipe network to reduce the slip and drag of its tracks. Chen Jun et al.\cite{jun2004study}, PAROYS-II\cite{park2010normal}, MR INSPECT-IV\cite{roh2005differential}, MR INSPECT-VI \cite{kim2013pipe,yang2014novel} are similar three-module pipe climbers that operates using different drive-types (refer Table I, comparing their functionalities with our proposed robot). Pre-determining the speeds of the tracks limit the robot to navigate bends only in orientations for which the speeds are preset \cite{vadapalli2019modular,suryavanshi2020omnidirectional,roh2005differential}. Thus, the need for planning the locomotion limits the robot’s success to only known environments.
\begin{figure}[ht!]
\centering
{\footnotesize TABLE I: In-pipe climbing robots with various drive mechanisms }
\includegraphics[width=3.4in]{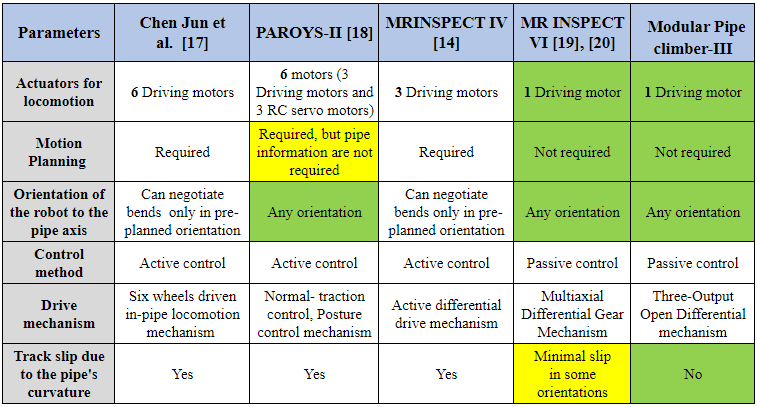}
\label{Figurex}
\vspace{-0.4in}
\end{figure}

\begin{figure}[ht!]
\centering
\includegraphics[width=3in]{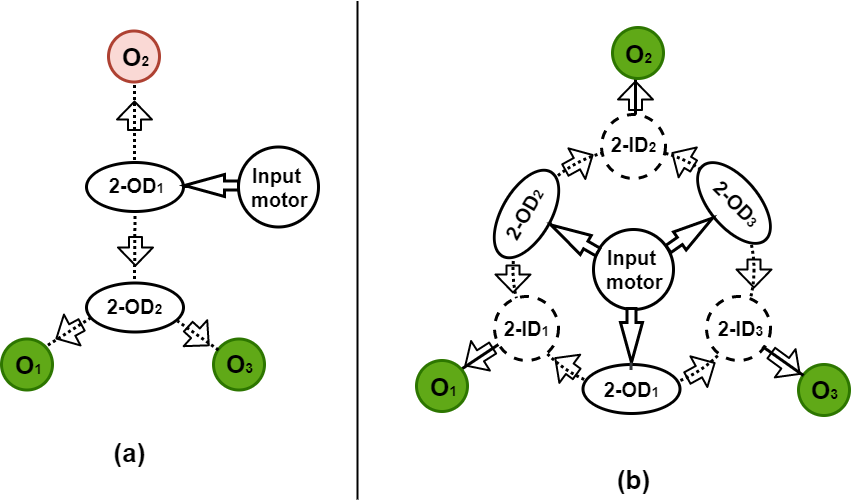}
\caption{\footnotesize (a) Schematics of conventional $3-OD$s (b) Schematics of the proposed $3-OOD$ mechanism}
\label{Figurea}
\vspace{-0.2in}
\end{figure}

The mentioned limitation can be addressed by using a differential mechanism which equips the robot to passively operate its tracks with differential speed. One such differential, the multi-axle differential gear, was implemented in the in-pipe robot MRINSPECT VI \cite{kim2013pipe,yang2014novel}. Incorporating a differential mechanism enabled MRINSPECT VI to considerably reduce the slip and drag of its wheels. However, the gear arrangement used in the differential is such that the mechanism favours one of its outputs $($output $O_{2})$ over the other two outputs $($output $O_1$ and output $O_3)$ as seen in the schematic Fig.~\ref{Figurea}(a) \cite{kim2013pipe}. As a consequence, when the robot traverses in pipes, one of the tracks moves faster than the other two causing slip or drag in a few orientations of the robot \cite{kim2016novel}. This limitation is transpired because all three outputs of the differential do not have equivalent dynamics with the input, Fig.~\ref{Figurea}(a) \cite{kim2013pipe,kim2016novel}. Solutions for three-output differentials ($3-OD$s) were also presented by S. Kota and S. Bidare $(1997)$ \cite{kota1997systematic} and Diego Ospina and Alejandro Ramirez-Serrano $(2020)$ \cite{ospina2020sensorless}. The differentials they proposed have schematics similar to Fig.~\ref{Figurea}(a) and exhibit similar limitations as the multi-axle differential gear \cite{kota1997systematic,ospina2020sensorless}. Due to the limitations present in the currently existing solutions, there exists the need for a three-module robot that traverses in-pipe without any slip or drag.


\textbf{Contribution:} We propose Modular Pipe Climber III, third in the series of our three-tracked pipe climbing robots. Our solution entails incorporating a differential mechanism to eliminate the slip and drag caused in robot's tracks due to the changing cross-section of the pipe while negotiating bends. Resultantly, we make two key contributions. Firstly, design of the Three-Output Open Differential $(3-OOD)$ \cite{vadapalli2021design}. $(3-OOD)$ is the first differential with all the three outputs sharing an equivalent kinematics relations to the input, as shown in the Fig.~\ref{Figurea}(b). As a result, $3-OOD$ has the novel ability to translate equal speeds and torque to all its outputs that are under equal loads, as theoretically represented in equations \eqref{27Velocity_O1=O2=O3} and \eqref{28Torque_O1=O2=O3}. Secondly, incorporating the $3-OOD$ enhances the robot’s ability to traverse pipe-networks without any slip or drag in any robot-orientation.

The robot is designed to traverse inside pipes of diameter $250$ mm to $280$ mm without requiring any active control. The paper discusses in detail the design of the robot and the novel 3-OOD mechanism it incorporates. The kinematics of 3-OOD mechanism and the robot is formulated and the robot's pipe navigating abilities are validated through experiments.
\vspace{-0.1in}
\begin{figure}[ht!]
\centering
\includegraphics[width=2.75in]{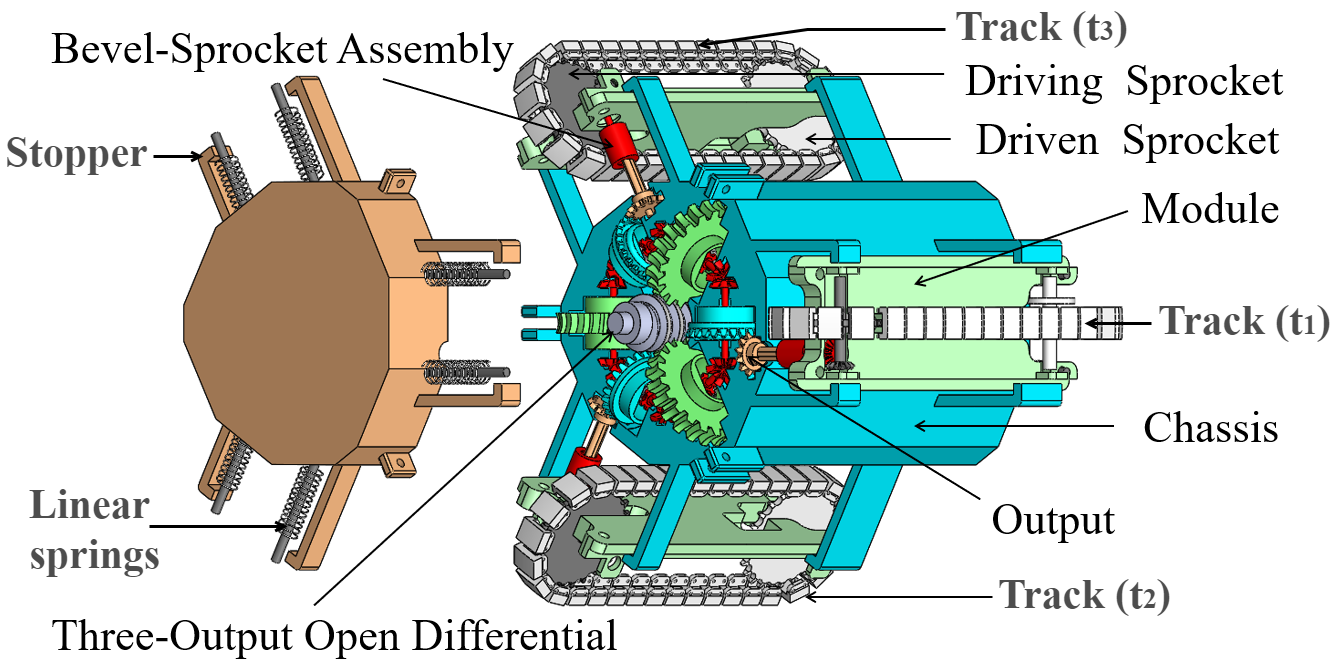}
\caption{\footnotesize Modular Pipe Climber III $($Exploded View$)$}
\label{Figure3}
\end{figure}
\vspace{-0.3in}

\section{DESIGN OF THE MODULAR PIPE CLIMBER III}
The Modular Pipe Climber III comprises of three tracks ($t1$, $t2$ and $t3$) that are run by a single motor via the $3-OOD$, as seen in Fig.~\ref{Figure3}. Tracks are housed on separate modules which are fixed 120$^\circ$ apart from each other on a nonagon-shaped center chassis, as shown in Fig.~\ref{Figure4}(a). The modules are connected to the chassis via wall-clamp mechanism which push the tracks against the inner wall of the pipe and provide traction, as illustrated in Fig.~\ref{Figure4}(a). The robot measures $200$ mm in length and $280$ mm in diameter.

The tracks consists of lugs inter-connected by chains that is rotated by the sprockets. The design of the tracks rely on the two ends of the sprocket for generating the tractive force. Therefore, when the robot immediately enter the bends from straight pipe sections, the driving sprocket and the driven sprocket in the track will have a relative difference in their angular velocities. To resolve this, each track is provided a small slack. The minimal slack helps in adjusting the tension of the track when there is a difference in the angular velocity between the two ends of the sprocket in the same module. Therefore, the tracks are constrained near the bends and the tractive force is always maintained between the pipe and the track during bends and straight pipes. The robot’s ability to operate in pipes is chiefly attributed to two mechanisms, the wall-clamp mechanism and the $3-OOD$.

\subsection{Wall-Clamp Mechanism}
\vspace{-0.1in}
\begin{figure}[ht!]
\centering
\includegraphics[width=3.3in]{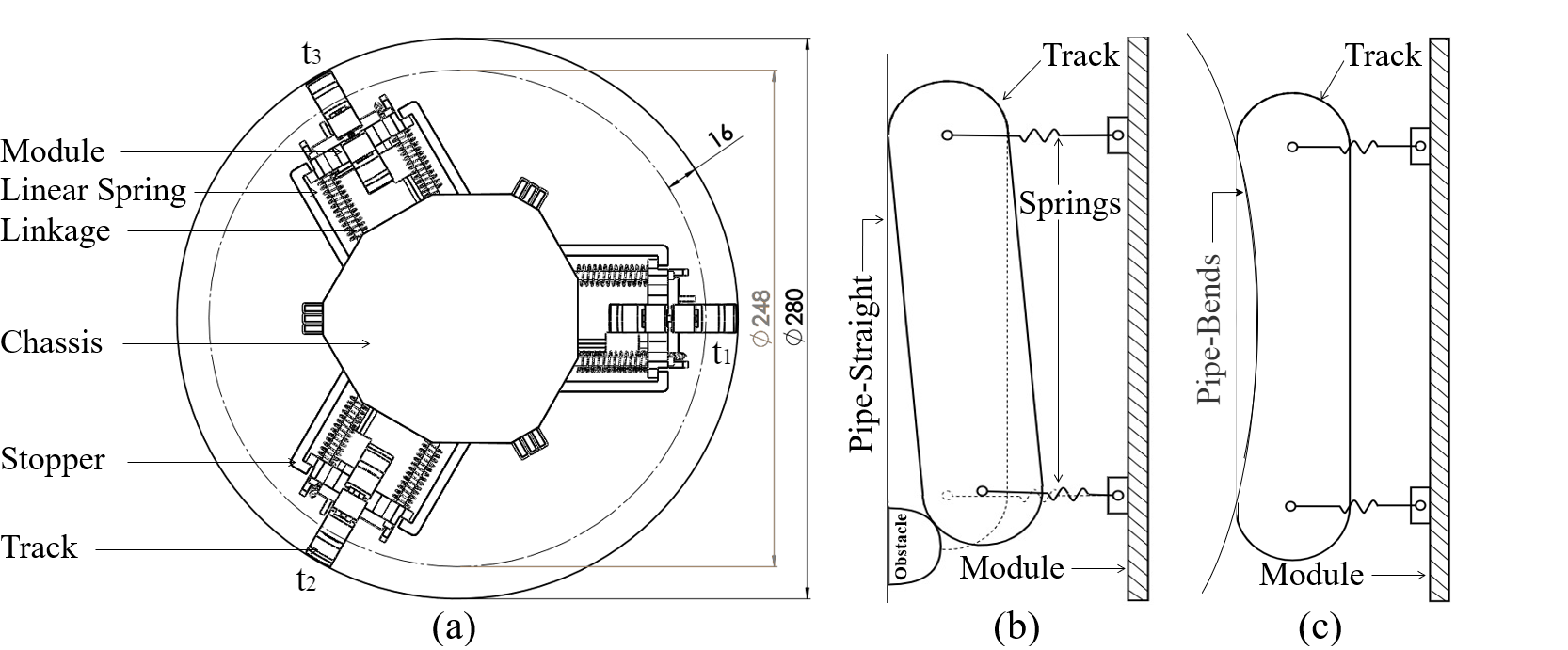}
\caption{\footnotesize Wall-Clamp Mechanism $($a$)$ Modular Pipe Climber III $($Front View$)$ $($b$)$ Asymmetrical compression in straight pipe $($c$)$ Navigation in Pipe-Bend}
\label{Figure4}
\end{figure}
\vspace{-0.1in}
The internal wall-clamp mechanism acts independently on each module as a compliant system. It comprises of linear springs, linkages and stopper as shown in Fig.~\ref{Figure4}(a). It is positioned between the chassis and the modules to provide radial compliance to the robot. Each module is spring loaded and connected to chassis by four linkages. The modules have the provision to slide along the linkages. Each linkage houses a linear spring which is pre-loaded and pushes the module radially outwards. The stoppers ensure that the modules are not pushed beyond the limit. The compliance enables the robot to passively vary its diameter by $32$ mm, from $280$ mm to $248$ mm, to maintain traction and adapt to various conditions the robot might experience inside a pipe. In addition, it allows each modules to compress asymmetrically to traverse over obstacles, as illustrated in Fig.~\ref{Figure4}(b).
\subsection{Three-Output Open Differential $(3-OOD)$}
\begin{figure}[ht!]
\centering
\includegraphics[width=2.7in]{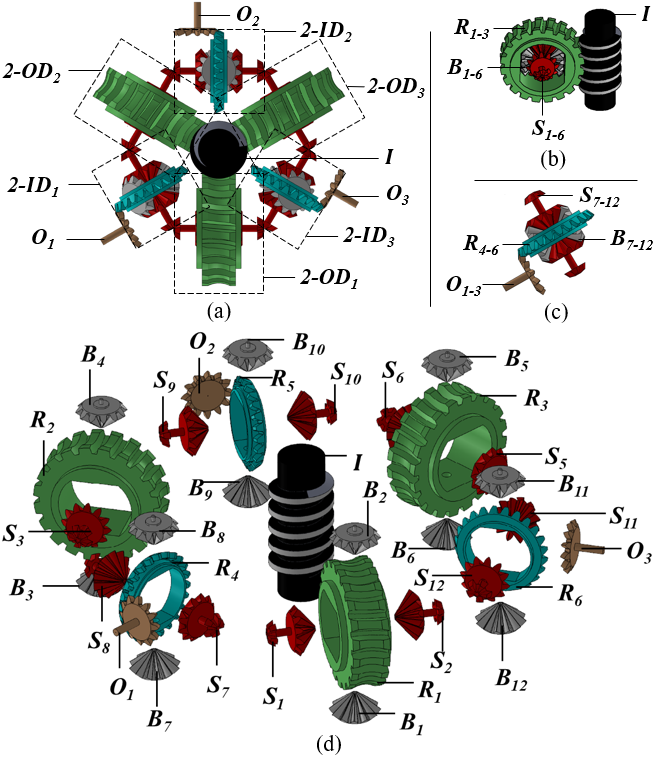}
\caption{\footnotesize $($a$)$ Three-Output Open Differential $($Top View$)$ $($b$)$ Two-Output Differential $(2-OD_{1-3})$ $($c$)$ Two-Input Differential $(2-ID_{1-3})$ $($d$)$ Three-Output Open Differential $($Exploded View$)$}
\label{Figure6}
\vspace{-0.3in}
\end{figure}

The $3-OOD$ is a novel element of the Modular Pipe Climber III. The differential is fitted inside the chassis and is connected to the three tracks via bevel-sprocket arrangement, shown in Fig.~\ref{Figure3}. Using the differential to drive the robot potentially eliminates the slip and drag of its tracks. This considerably reduces the stresses on the robot and thus, providing a smoother motion to the robot. The $3-OOD$ comprises of a single input $(I)$, three two-output open differentials $(2-OD_{1-3})$, three two-input open differentials $(2-ID_{1-3})$ and three outputs $(O_{1-3})$, as shown in Fig.~\ref{Figure6}. The differential’s input is located at its centre and the three two-output differentials $(2-OD_{1-3})$ are arranged around the input with an angle of $120{^\circ}$ between them. The two-input differentials $(2-ID_{1-3})$ are fitted symmetrically between the two-output differentials. The single output of each of the three two-input differentials $(2-ID_{1-3})$ form the three outputs $(O_{1-3})$ of the $3-OOD$, as seen in Fig.~\ref{Figure6}.
\vspace{0.1in}

The input $(I)$ from the worm gear simultaneously provides motion to the three two-output differentials $(2-OD_{1-3})$ which further translates the motion to its neighbouring two-input differentials $(2-ID_{1-3})$ depending on the load their side gears $(S_{1-6})$ experience, Fig.~\ref{Figure6}(a). The $(2-OD_{1-3})$ translate differential speed to its neighbouring $(2-ID_{1-3})$ if its two side gears $(S_{1-6})$ operate under different loads. The motion received by the two side gears $(S_{7-12})$ of the respective two-input differentials $(2-ID_{1-3})$ is translated to the three outputs $(O_{1-3})$ Fig.~\ref{Figure6}. When the two side gears of a two-input differential receive different speeds, it translates the differential speed to a single output. The six differentials $(2-OD_{1-3})$ and $(2-ID_{1-3})$ work together to translate motion from the input $(I)$ to the three outputs $(O_{1-3})$.

\textbf{\small Novelty of \textit{3-OOD}:\normalsize} The three outputs of the $3-OOD$ have equivalent input to output dynamics which can be noticed in its schematic in Fig.~\ref{Figurea}(b). Additionally, the outputs share the same dynamics with each other. As a result, the change in loads for one of the outputs will have an equal effect on the other two outputs. Thus, the $3-OOD$ achieves the novel result of operating its three outputs with differential speed when they are under varied loads and with equal speeds when the outputs are subjected to equivalent loads, as theoretically derived in the equations \eqref{27Velocity_O1=O2=O3} and \eqref{28Torque_O1=O2=O3}. For instance, when the outer module operates at a different speed and the other two inner modules are under equivalent loads in the bends, then both the inner modules operate with equal angular velocities and equal torques. This is another novel result realised by using the $3-OOD$ in the pipe climber. Owing to these results, the $3-OOD$ is the first three-output differential whose functioning is analogous to that of the traditional two-output open differential.

This $3-OOD$ is specifically designed to be used inside the Modular Pipe Climber III. It equips the robot with differential speed inside a pipe-bend so that the track travelling the longer distance rotates faster than the track travelling the shorter distance, but when moving inside a straight pipe section the three tracks rotate with equivalent speeds.
\section{MATHEMATICAL MODEL}
\subsection{Angular velocity and torque outputs of the $3-OOD$}
This section presents the angular velocity and torque relation between the outputs $(O_{1-3})$ and the input ($I$). The kinematics and the dynamics of the $3-OOD$ are derived by the means of the bond graph modelling technique \cite{vadapalli2021design}, showing the novel ability of the mechanism to exhibit equivalent output to input angular velocity and torque.

By equating the input to the side gear equation and the output to the side gears relation, we get the input to output relation for the angular velocity, refer \cite{vadapalli2021design}.

\begin{equation}
\begin{split}
& \hspace{-0.06in} \omega _{O1} \hspace{-0.04in}=\hspace{-0.04in} \frac{2j(\omega_i)}{k} \hspace{-0.04in}-\hspace{-0.04in} \frac{j(\omega _2+\omega _4)}{2},\hspace{0.025in}
\omega _{O2} \hspace{-0.04in}=\hspace{-0.04in} \frac{2j(\omega_i)}{k} \hspace{-0.04in}-\hspace{-0.04in} \frac{j(\omega _3+\omega _5)}{2},\\ &
\hspace{0.7in} \omega _{O3} \hspace{-0.04in}=\hspace{-0.04in} \frac{2j(\omega _i)}{k} \hspace{-0.04in}-\hspace{-0.04in} \frac{j(\omega _1+\omega _6)}{2}
\end{split}
\label{17velocity_O3-I}
\end{equation}

where ($1/k$= 1/20) is the gear ratio of the input to the ring gears ($R_{1-3}$), while (j = 2/1) is the gear ratio of the ring gears ($R_{4-6}$) to the outputs. $\omega_i$ and $\tau_i$ are the angular velocity and torque of the input.

Similarly, output to the input torque relations are obtained 

\begin{equation}
\begin{split}
& \tau_{O1}=\frac{k(\tau_i)}{3j}-\frac{(I_1\Dot{\omega}_7+I_3\Dot{\omega}_8)}{j}, \ \hspace{0.2in}  \tau_{O2}=\frac{k(\tau_i)}{3j}\hspace{0.025in}- \\ &\frac{(I_4\Dot{\omega}_9+I_6\Dot{\omega}_{10})}{j}, \ \hspace{0.2in}  \tau_{O3}=\frac{k(\tau_i)}{3j}-\frac{(I_2\Dot{\omega} _{12}+I_5\Dot{\omega}_{11})}{j}
\label{23torque_O3-I}
\end{split}
\end{equation}

\textbf{\small Equal speeds and torque:\normalsize} Equations \eqref{17velocity_O3-I} and \eqref{23torque_O3-I} show that all the three outputs of the differential share equivalent angular velocity and torque relations with the input. Since the side gears are all identical, they exhibit equal inertia $(I_1=I_2=I_3=I_4=I_5=I_6)$. When all three outputs $(O_{1-3})$ are under equal loads (resistive torque `$\tau_R$'), 
$\omega _n= \frac{\omega _i}{k}$, where $n$ goes from $1$ to $12$. Substituting these relations in \eqref{17velocity_O3-I} and \eqref{23torque_O3-I},

\begin{equation}
{\bf\omega _{O1}}={\bf\omega _{O2}}={\bf\omega _{O3}}=\frac{j(\omega _i)}{k},
\label{27Velocity_O1=O2=O3}
\end{equation}

\begin{equation}
{\bf\tau _{O1}}={\bf\tau _{O2}}={\bf\tau _{O3}}=\frac{k(\tau_i)}{3j} - \frac{2(I_1\dot{\omega}_1)}{j} - \tau_R,
\label{28Torque_O1=O2=O3}
\end{equation}
Equations \eqref{27Velocity_O1=O2=O3} and \eqref{28Torque_O1=O2=O3} illustrate the novel ability of the $3-OOD$ to translate equal angular velocities and torque to all its three outputs that are unconstrained $(\tau_R = 0)$ or under equal loads. When any one of the outputs ($O_1$) is constrained due to a load $(\tau_R)$, then the other two outputs ($O_2$ and $O_2$) behave according to the variation of $O_1$ due to $(\tau_R)$.
\subsection{Kinematics Of The Robot}
The angular velocities $\omega_{O1}$, $\omega_{O2}$ and $\omega_{O3}$ of the three outputs $(O_1$, $O_2$ and $O_3)$ translate as linear velocities (mm/s) --- $v_{t1}$, $v_{t2}$ and  $v_{t3}$ of the three tracks (${t1}$, ${t2}$, ${t3}$)

\begin{figure}[ht!]
\centering
\hspace*{-0.5cm}
\includegraphics[width=3.3in]{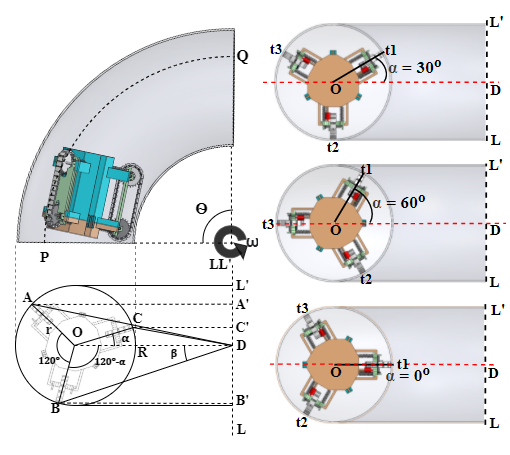}
\vspace{-0.2in}
\caption{\footnotesize Planar Representation  of  the  Robot’s  Navigation in Pipe-Bends for different orientations of the robot 
}
\label{Figure9}
\vspace{-0.2in}
\end{figure}

\begin{equation}
\begin{split}
v_{t1} \hspace{-0.02in}=\hspace{-0.02in} \frac{\omega_{O1} \hspace{-0.02in}\times\hspace{-0.02in} \pi d}{60}, \quad v_{t2} \hspace{-0.02in}=\hspace{-0.02in} \frac{\omega_{O2} \hspace{-0.02in}\times\hspace{-0.02in} \pi d}{60}, \quad v_{t3} \hspace{-0.02in}=\hspace{-0.02in} \frac{\omega_{O3} \hspace{-0.02in}\times\hspace{-0.02in} \pi d}{60}
\label{9velovity_T-O}
\end{split}
\end{equation}

where $(d = 80$ mm$)$ is the diameter of the tracks. Figure~\ref{Figure9} shows the planar representation of the robot’s navigation inside a pipe-bend of angle $\theta$. The center of the pipe is marked $O$ and $R$ denotes the radius of curvature of the pipe-bend about the axis $LL'$. $A$, $B$ and $C$ are the points of contact of the three tracks ($t1$, $t2$, $t3$) of the robot with the inner wall of the pipe, with $r$ being the radius of the pipe. The robot’s center matches the pipe's center $O$ and $AA'$, $BB'$ and $CC'$ are the perpendicular distance between the tracks of the robot and the axis $LL'$. Pipe-bends are usually designed to have a constant radius of curvature \cite{gross1953flexibility}. Hence, as the robot navigates inside a pipe-bend, the paths traced by its three tracks have a uniform curvature, following which the distances $AA'$, $BB'$ and $CC'$ remain constant. The differential in the robot prompts the track that is farther away from $LL'$ to travel faster than the track that is closer to $LL'$. The three tracks of the robot are $120^\circ$ apart from each other. $\alpha$ is the angle subtended between the inner track of the robot to its radius of curvature of the pipe. It lies between the inner track of the robot $CO$ with $OD$, where $D$ is the orthogonal projection of the point $O$ on the axis $LL'$. In $\Delta OAD$, the angle between $OD$ and $DA$ is $\beta$  $(\angle ODA = \beta)$. Since the angles $(\angle AOD = 120^\circ- \alpha)$ and $(\angle COD = \alpha)$
\begin{equation}
(AD)^2 = {\textstyle r^2 + R^2 + 2Rr\times\cos({120^\circ-\alpha})} \label{eq1}
\end{equation}
\begin{equation}
\cos{\beta} = \frac{(AD)^2 + R^2 - r^2}{2\times R\times(AD)}
\label{eq2}
\end{equation}

The lines connecting $AA'$ and $OD$ are parallel and hence, the angles $\angle DAA'$ and $\angle ODA$ are equal ($\angle$ $D$ $AA'$ = $\angle ODA$ = $\beta$). Therefore, the radius of curvature $AA'$ of the path traced by the track at $A$

\vspace{-0.05in}
\begin{equation}
AA' = AD \times \cos{\beta}.
\label{eq3}
\end{equation}
When negotiating a pipe-bend of angle $\theta$, the total distance travelled by track $1$ at $A$ is given by $(D_{t1} = \theta \times$ $AA'$). The velocity of the track
\vspace{-0.05in}
\begin{equation}
V_{t1} = \frac{d}{dt}(D_{t1}) = \omega \times AA',
\label{eq5}
\end{equation}
where $\omega$ is the angular velocity of the robot inside a bend, Fig.~\ref{Figure9}. Similarly, the speeds of track 2, track 3 located respectively at $B$ and $C$
\begin{equation}
V_{t2} \hspace{-0.01in}=\hspace{-0.01in} \frac{d}{dt}(D_{t2}) \hspace{-0.01in}=\hspace{-0.01in} \omega \times BB', \ \hspace{0.01in} V_{t3} \hspace{-0.01in}=\hspace{-0.01in} \frac{d}{dt}(D_{t3}) \hspace{-0.01in}=\hspace{-0.01in} \omega \times CC'. \label{eq6}
\end{equation}
Using the values $r$ (radius of the pipe, $r$ $=$ $138$ mm) and $R$ (radius of curvature of the pipe-bend, $R$ $=$ $419$ mm), the speeds of the three tracks of the robot in any orientation $\alpha$ inside any pipe-bend of angle $\theta$ can be calculated.
\begin{table}[t]
{\footnotesize TABLE II: Comparison of Theoretical and Experimental Results In Different Orientations $(\alpha)$ of the Robot and Estimation of Slip}
\begin{center}
  \centering
  \scalebox{0.745}{
  \begin{tabular}{ | m{5em} | m{2.4cm} | m{2.6cm} | m{1.5cm} | m{1.35cm} | }
    \hline
    Orientation & Theoretical Distance & Experimental Distance & \hspace{0.15in} Error & $|$Error$|$ $<$ $LC_{error}$ \\ \hline
    \begin{minipage}{1cm}
    \centering
    \vspace{0.03in}
      \includegraphics[width=15mm]{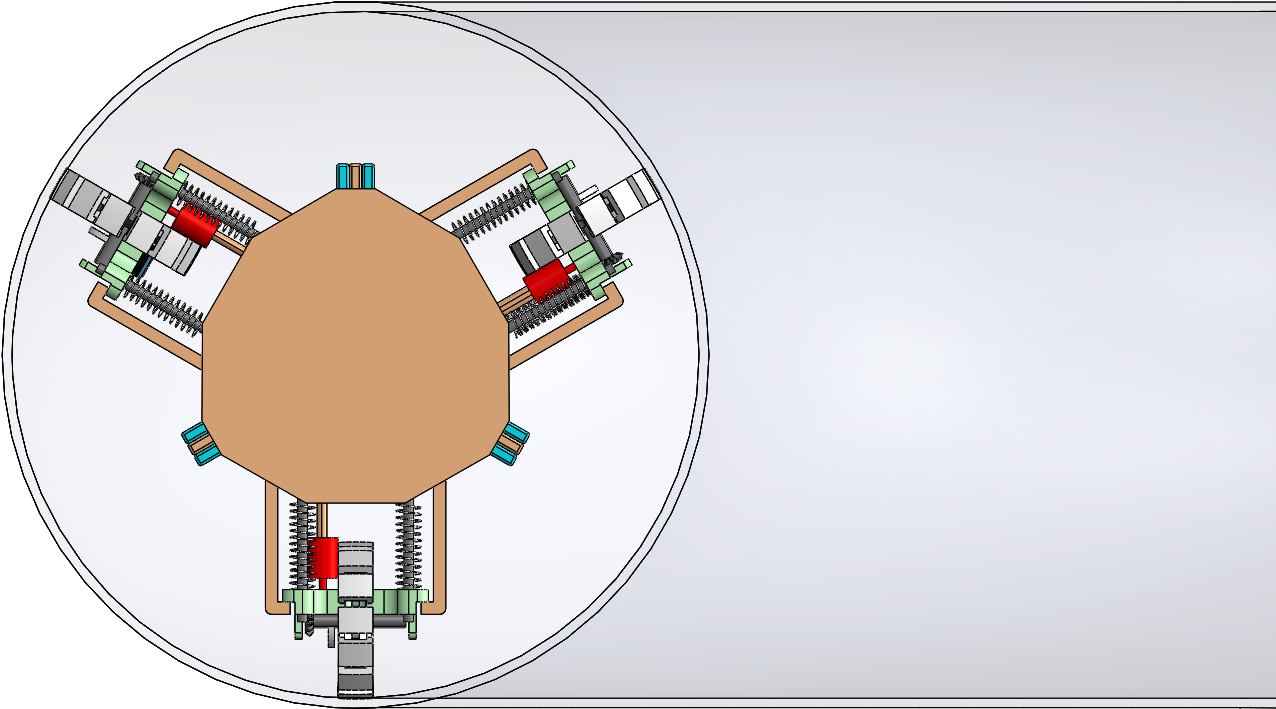}
      \caption*{\footnotesize $\alpha$ = $30^\circ$}
    \end{minipage}
    &
    \makecell{
      $D_{t1}$ = 2,453.29 mm \\
      $D_{t2}$ = 3,016.49 mm \\
      $D_{t3}$ = 3,579.69 mm \\
      $D_{R}$ = 3,016.49 mm}
    & 
      \makecell{
      $D_{t1}$ = 2,412.70 mm \\
      $D_{t2}$ = 3,015.80 mm \\
      $D_{t3}$ = 3,568.70 mm \\
      $D_{R}$ = 2,999.07 mm}
    &
    \makecell{
      -40.49 mm \\
      -0.69 mm \\
      -10.99 mm \\
      -17.42 mm}
    &
    \makecell{
      Yes \\
      Yes \\
      Yes \\
      Yes}
    \\ \hline
    \begin{minipage}{1cm}
    \centering
    \vspace{0.03in}
      \includegraphics[width=15mm]{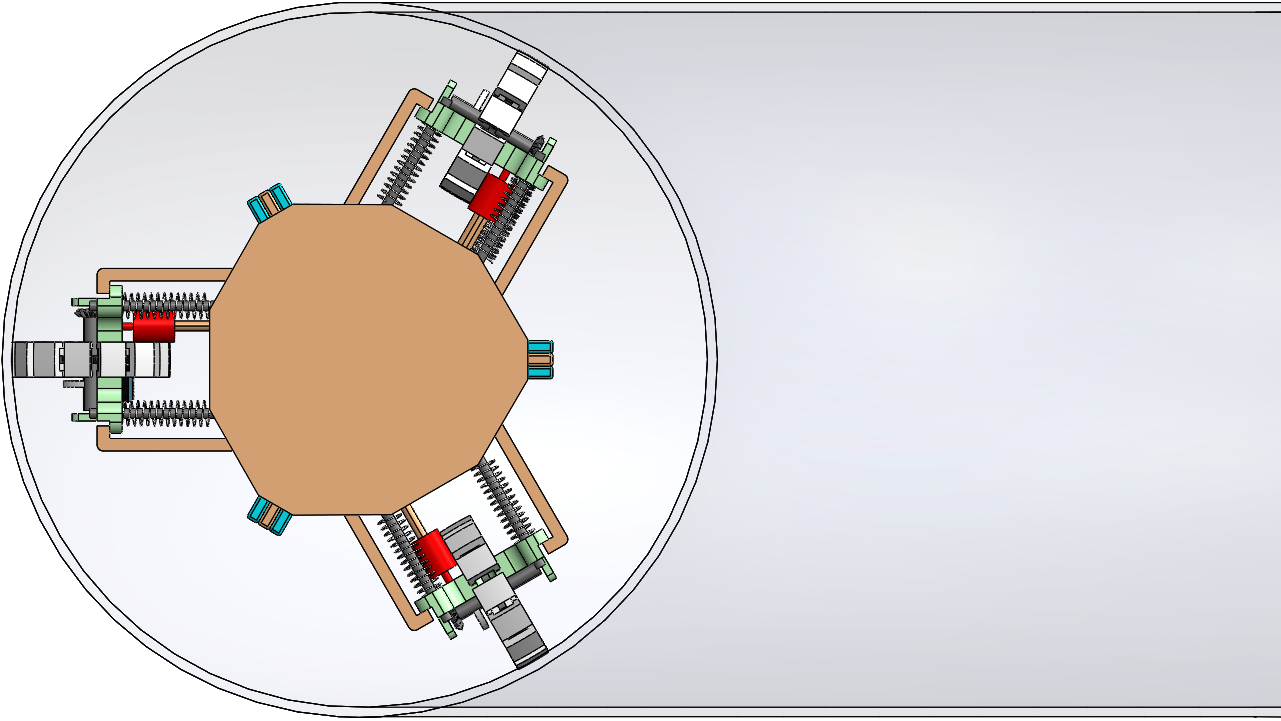}
      \caption*{\footnotesize $\alpha$ = $60^\circ$}
    \end{minipage}
    &
    \makecell{
      $D_{t1}$ = 2,691.34 mm \\
      $D_{t2}$ = 2,691.34 mm \\
      $D_{t3}$ = 3,666.80 mm \\
      $D_{R}$ = 3,016.49 mm}
    & 
    \makecell{
      $D_{t1}$ = 2,664.00 mm \\
      $D_{t2}$ = 2,714.30 mm \\
      $D_{t3}$ = 3,619.00 mm \\
      $D_{R}$ = 2,999.10 mm}
    &
    \makecell{
      -27.34 mm \\
      +22.96 mm \\
      -47.80 mm \\
      -17.39 mm}
    &
    \makecell{
      Yes \\
      Yes \\
      Yes \\
      Yes}
    \\ \hline
      \begin{minipage}{1cm}
    \centering
    \vspace{0.03in}
      \includegraphics[width=15mm]{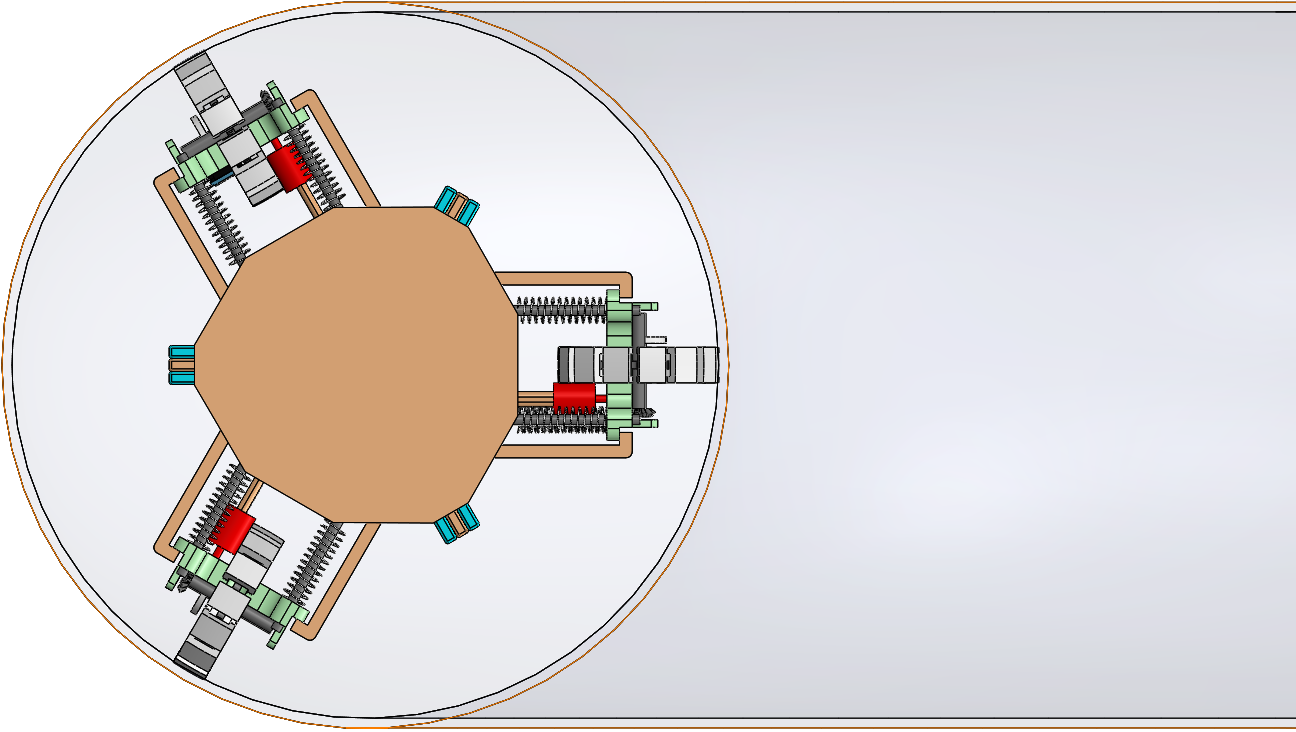}
      \caption*{\footnotesize $\alpha$ = $0^\circ$}
    \end{minipage}
    &
    \makecell{
      $D_{t1}$ = 2,366.18 mm \\
      $D_{t2}$ = 3,341.65 mm \\
      $D_{t3}$ = 3,341.65 mm \\
      $D_{R}$ = 3,016.49 mm}
    & 
    \makecell{
      $D_{t1}$ = 2,412.70 mm \\
      $D_{t2}$ = 3,317.40 mm \\
      $D_{t3}$ = 3,367.70 mm \\
      $D_{R}$ = 3,032.60 mm} 
    &
    \makecell{
      +46.52 mm \\
      -24.25 mm \\
      +26.05 mm \\
      +16.11 mm}
    &
    \makecell{
      Yes \\
      Yes \\
      Yes \\
      Yes}
    \\ \hline  
    \end{tabular}}
\end{center}
\vspace{-0.3in}
\end{table}

The speed of the robot remains constant in all orientations and is always equal to the average of the speeds of the three tracks, Fig.~\ref{Figurezo}. From the distance travelled by the robot and the input velocity, the time taken shown in Fig.~\ref{Figurez} to negotiate the pipe is calculated. The formulated linear speeds for each tracks at different orientations ($\alpha$ = $30^\circ$, $\alpha$ = $60^\circ$, $\alpha$ = $0^\circ$) were used to attain the theoretical distance.

\begin{figure}[ht!]
\centering
\includegraphics[width=3in]{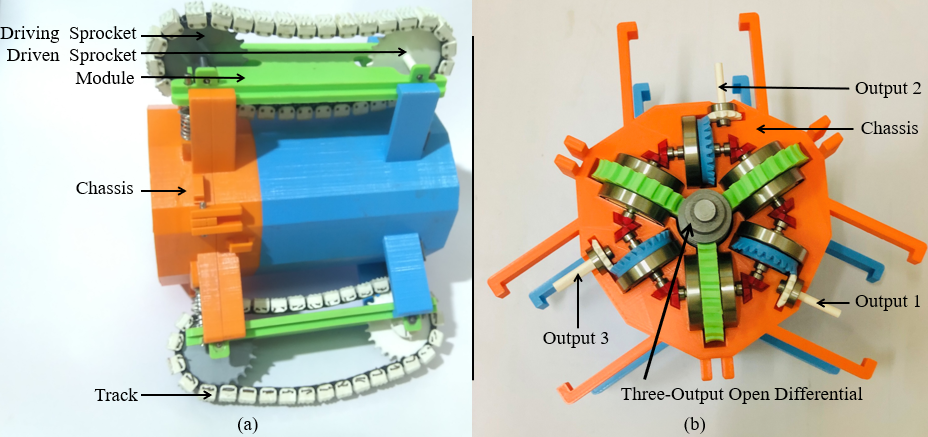}
\caption{\footnotesize Prototype $($a$)$ Modular Pipe Climber III $($b$)$ The $3-OOD$}
\label{Figure11}
\end{figure} 
\vspace{-0.2in}

\begin{figure*}[ht!]
\centering
\includegraphics[width=7in]{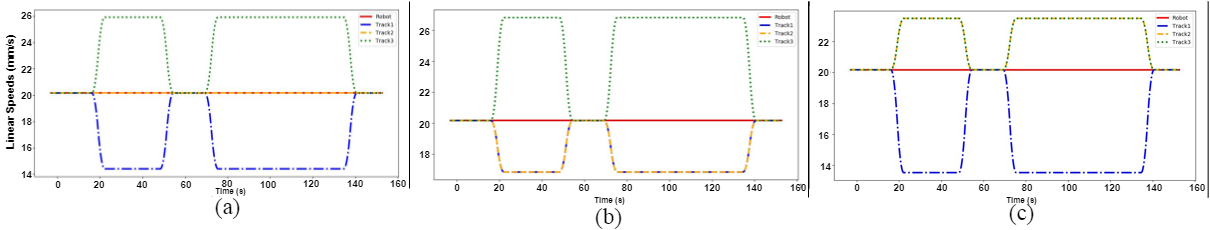}
\caption{\footnotesize Theoretical results for $($a$)$ $\alpha$ = $30^\circ$ $($b$)$ $\alpha$ = $60^\circ$ $($c$)$ $\alpha$ = $0^\circ$, where the robot traverses in vertical pipe $(0$ - $20.9$ s$)$, $90^\circ$ bend $(20.9$ - $53.6$ s$)$, horizontal pipe $(53.6$ - $74.2$ s$)$ and $180^\circ$ bend $(74.2$ - $139.7$ s$)$}
\label{Figurezo}
\vspace{-0.1in}
\end{figure*}

\begin{figure*}[ht!]
\centering
\includegraphics[width=7in]{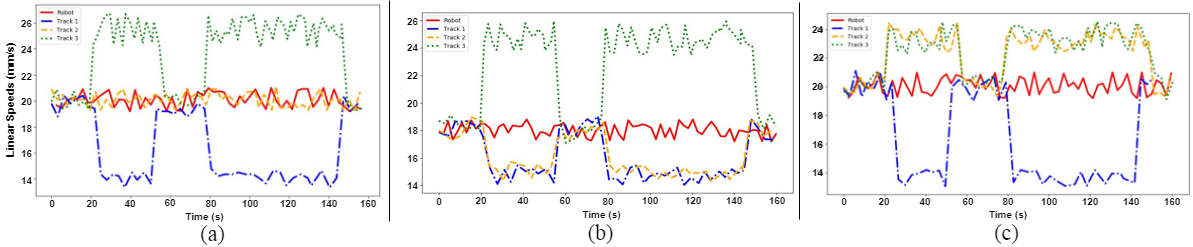}
\caption{\footnotesize Experimental results for $($a$)$ $\alpha$ = $30^\circ$ $($b$)$ $\alpha$ = $60^\circ$ $($c$)$ $\alpha$ = $0^\circ$, where the robot traverses in vertical pipe from $(0$ - $20$ s$)$, $90^\circ$ bend $(20$ - $56$ s$)$, horizontal pipe $(56$ - $75$ s$)$ and $180^\circ$ bend $(75$ - $147$ s$)$}
\label{Figurez}
\vspace{-0.2in}
\end{figure*}

\section{EXPERIMENTATION ON THE PROTOTYPE}

\begin{figure}[ht!]
\centering
\includegraphics[width=2.3in]{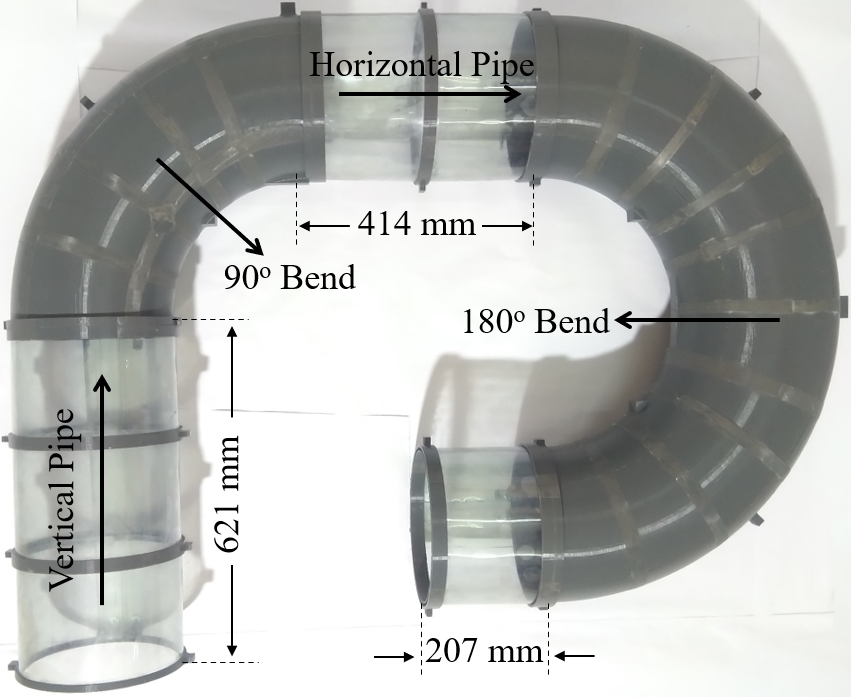}
\caption{\footnotesize Experimental Pipe-Network}
\label{Figure12}
\end{figure}

\begin{figure}[ht!]
\centering
\includegraphics[width=3in]{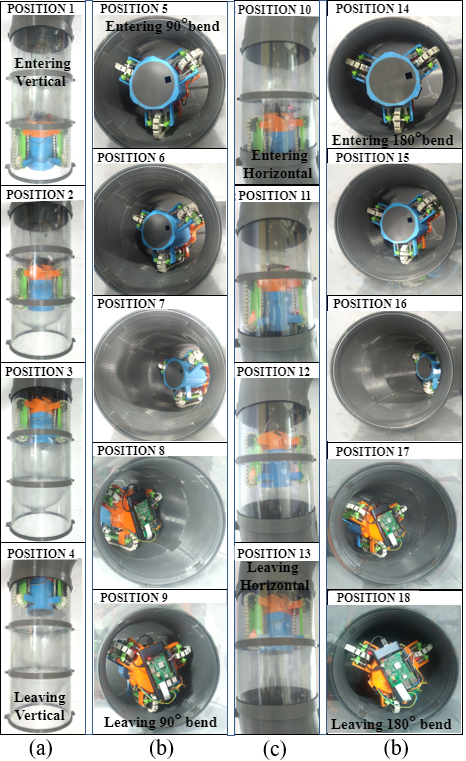}
\caption{\footnotesize Robot's Traversal in $($a$)$ Vertical Pipe $($b$)$ 90$^\circ$ Bend $($c$)$ Horizontal Pipe $($d$)$ 180$^\circ$ Bend}
\label{Figure14}
\vspace{-0.25in}
\end{figure}

A prototype of the Modular Pipe Climber III, as shown in Fig.~\ref{Figure11}, is built to further validate the robot’s ability to traverse pipe-networks. The differential of the prototype is powered by a PDS4360-12-60 DC-motor which produces 1.177 N-m nominal torque at $48$ rpm. The robot is controlled remotely via Raspberry Pi 3 Model B+. Optical encoders are installed on the robot to measure the speeds of the three tracks. The encoders' sensors record a reading for every $50.26$ mm which is taken as the least count error for the experiment $({\bf LC_{error}} = 50.26$ mm$)$. The robot is tested inside a pipe-network built to the dimensions of the simulated environment, constructed as per the ASME B16.9 standard NPS 11 and Schedule 40, as shown in Fig.~\ref{Figure12}.
The speeds of the three tracks and the robot in cases $(a)$ $\alpha $ = $ 30^\circ$, $(b)$ $\alpha $ = $ 60^\circ$ and $(c)$ $\alpha $ = $ 0^\circ$ are as shown in Fig.~\ref{Figurez}. The robot's navigation in different sections of the pipe with the robot's orientation being $\alpha $ = $ 30^\circ$ is shown in Figure~\ref{Figure14}. When traversing in vertical pipe $(0$ $-$ $20$ s$)$ and horizontal pipe $(56$ $-$ $75$ s$)$, the observed mean linear speeds of the three tracks and the robot are $\mu_{vt1}$ = $19.5$ mm/s, $\mu_{vt2}$ = $19.9$ mm/s, $\mu_{vt3}$ = $19.9$ mm/s and $\mu_{vR}$ = $19.7$ mm/s. Inferring from Fig.~\ref{Figurez}, the speeds of the three tracks and the robot in the pipe-straights are approximately equal with an error of $v_{error}$ = $0.4$ mm/s. When negotiating the 90$^\circ$ bend $(20$ - $56$ s$)$ and 180$^\circ$ bend $(75$ - $147$ s$)$, the mean linear speed of the track, $t1$ ($\mu_{vt1}$ = $14.1$ mm/s) travels the shortest distance than the mean linear speeds of track $t2$, $(\mu_{vt2}$ = $19.5$ mm/s$)$ and track, $t3$ $(\mu_{vt3}$ = $25.1$ mm/s$)$. At orientation ($\alpha$ = $30^\circ$) shown in Figure~\ref{Figurez}(a), we can deduce that the track $t3$, which travels the longest distance, moves the fastest among the three tracks. The robot takes $153.9$ seconds to navigate the pipe-network. The total distance travelled by the three tracks and the robot in the orientation $\alpha $ = $ 30^\circ$ are $D_{t1}$ = $2,412.7$ mm, $D_{t2}$ = $3,015.8$ mm, $D_{t3}$ = $3,568.7$ mm and $D_{R}$ = $2,999.07$ mm respectively, as shown in Table II.

\textbf{\small Navigation without slip and drag:\normalsize} The distances recorded experimentally match the theoretical results as seen in Table II, with the maximum error $({\bf D_{t1-error}}$ $=$ $-40.49$ mm$)$ occurring at track $t1$, which amounts to a percentage error of $\textbf{1.65\%}$. The recorded error ${\bf D_{t1-error}}$ $($distance calculated theoretically subtracted from the distance recorded experimentally, for track $t1$ is within the least count error estimated for the experiment $({\bf LC_{error}}$ $=$ $50.26$ mm$)$, as shown in Table II. The recorded errors in the distance traveled for the three tracks and the robot in all three cases $\alpha $ = $ 30^\circ$, $\alpha $ = $ 60^\circ$ and $\alpha $ = $ 0^\circ$ are within the estimated least count error with the highest deviation $({\bf D_{t3-error}}$ $=$ $-47.80$ mm$)$ occurring at track $t3$ in $\alpha $ = $ 60^\circ$ orientation, as illustrated in Table II. These results support our proposition that the robot traverses without any slip or drag. Furthermore, it is experimentally observed that the speeds of the three tracks and the robot, in all the three orientations ($\alpha $ = $ 30^\circ$, $\alpha $ = $ 60^\circ$ and $\alpha $ = $ 0^\circ$) in Fig.~\ref{Figurez}, show a strict resemblance with theoretical results, as seen in Fig.~\ref{Figurezo}. This further asserts the robots ability to navigate without slip and drag in any orientation of the robot. The Modular Pipe Climber III achieves this novel result because all the three outputs from the $3-OOD$ mechanism have equivalent angular velocity and torque distribution to each other and to the input. The three tracks operate at almost equal velocities when moving in pipe-straights. In the pipe bends, the track velocities modulate according to the radius of curvature of the pipe at any inserted orientation of the robot. The minute differences in the velocities can be attributed to the tolerances present in the design and the experimentation procedure. The pipe-climber is also successfully tested in other random orientations.

\section{CONCLUSION}
The Modular Pipe Climber III incorporating the novel Three-Output Open Differential is presented. The differential is designed such that its functioning ability is analogous to the traditional two-output open differential. The differential enables the robot to negotiate pipe-bends without the need for any active control and additional control efforts. The robot is tested in a complex pipe-network in multiple orientations and the results verify the robot's ability to navigate different pipe sections in all orientations without any slip or drag due to the changing cross-section of the pipe. Avoiding slip and drag considerably reduces the stresses experienced by the robot and provides a smoother locomotion for the pipe climber.


\vspace{-0.1in}

\bibliographystyle{./bibliography/IEEEtran} 
\bibliography{./bibliography/IEEEabrv,./bibliography/root}

\end{document}